\documentclass[10pt,twocolumn,letterpaper]{article}

\usepackage{cvpr}
\usepackage{times}
\usepackage{epsfig}
\usepackage{graphicx}
\usepackage{amsmath}
\usepackage{amssymb}
\usepackage{color}
\usepackage{times}
\usepackage{epsfig}
\usepackage{algorithm}
\usepackage{algorithmicx}
\usepackage{algpseudocode}
\usepackage{graphics}
\usepackage{threeparttable}
\usepackage{color}
\usepackage[normalem]{ulem}
\usepackage{multirow}
\usepackage{float}
\usepackage{amsfonts}
\usepackage{bm}
\usepackage{subfig}
\usepackage{array}
\usepackage[table]{xcolor}
\usepackage{colortbl}
\usepackage{pifont}
\usepackage{cite}
\usepackage{diagbox}
\usepackage{rotating}
\usepackage{booktabs}
\usepackage{overpic}
\usepackage{textcomp}
\usepackage{contour}
\usepackage{enumitem}

\usepackage[pagebackref=true,breaklinks=true,letterpaper=true,colorlinks,bookmarks=false]{hyperref}
\cvprfinalcopy 

\def\cvprPaperID{****} 

\ifcvprfinal\fi
\begin{document}

\title{Shed Various Lights on a Low-Light Image: \\ Multi-Level Enhancement Guided by Arbitrary References}

\author{Ya'nan Wang\footnotemark[1]\quad  Zhuqing Jiang\footnotemark[1]\quad  Chang Liu\quad  \\
	\quad Kai Li\quad  Aidong Men\quad Haiying Wang\quad\\
	\small Beijing University of Posts and Telecommunications \quad\\
	{\tt\small $\{$wynn,jiangzhuqing,chang\_liu,xiaoyao125656,menad,why$\}$@bupt.edu.cn}
}

\maketitle
\thispagestyle{empty}
\renewcommand{\thefootnote}{\fnsymbol{footnote}}
\footnotetext[1]{The first two authors contribute equally to this work.}

\vspace{-8pt}
\begin{abstract}
It is suggested that low-light image enhancement realizes one-to-many mapping since we have different definitions of NORMAL-light given application scenarios or users' aesthetic. However, most existing methods ignore subjectivity of the task, and simply produce one result with fixed brightness. This paper proposes a neural network for multi-level low-light image enhancement, which is user-friendly to meet various requirements by selecting different images as brightness reference. Inspired by style transfer, our method decomposes an image into two low-coupling feature components in the latent space, which allows the concatenation feasibility of the content components from low-light images and the luminance components from reference images. In such a way, the network learns to extract scene-invariant and brightness-specific information from a set of image pairs instead of learning brightness differences. Moreover, information except for the brightness is preserved to the greatest extent to alleviate color distortion. Extensive results show strong capacity and superiority of our network against existing methods.
\end{abstract}

\vspace{-6pt}
\section{Introduction}
Nowadays, taking photos is convenient with omnipresence of cameras on multiple devices. However, photos often suffer degradations due to the environment and equipment limitations, such as low contrast, noise, and color distortion. Since vision perception is related to application scenarios and users' aesthetic, image enhancement should be guided by these factors to improve quality of photos. Although existing professional software provides tools for manipulating photos to help users get their visually pleasing images, these tools are either user-unfriendly or working inferior. Thus, a low-light image enhancement method that meets different needs is essential.

With rapid development of deep learning, various methods have been proposed to enhance low-light images. Recent algorithms in~\cite{Lore2017,2017MSR,Chen2018,Chenchen2018,Zamir2020MIRNet,guo2020zero,jiang2019enlightengan,Wang2019} enhance low-light images to a fixed brightness, that is, the algorithms learn brightness difference of training data pairs. Thus, they are inflexible and enhance images without diversity. Such methods ignore the subjectivity. In~\cite{zhang2019kindling}, the light level is adjusted by a strength ratio, but it may not be an wieldy descriptor for users since the relationship between the perceived light level and the strength ratio is non-linear. \cite{kim2020pienet} models user preferences as vectors to guide the enhancement process, yet the preparation of preference vectors is complicated. Furthermore, except for brightness information, color information is also learned in the vector, which leads to color distortion.

In this paper, we propose a deep learning algorithm for multi-level low-light image enhancement guided by arbitrary images as brightness references. Inspired by style transfer, we assume that an image consists of a content component and a luminance component in the latent space, which is proved to be reasonable in our experiments. Specifically, content components refer to scene-invariant information during the enhancement, while luminance components represent brightness-specific information.

A similar but nontrivial theory is Retinex\cite{Land1977}, which decomposes an image into two sub-images, namely reflectance and illumination. It enhances low-light images by adjusting the illumination, and then recombine it with corresponding reflectance. In contrast, our feature components are low-coupling, which allows a new image generated by concatenating two feature components from different images.

Our main contributions are summarized as follows:

\noindent 1)
The proposed network decomposes images into content components and luminance components in the latent space, which are independent of each other. The feature components of different images are concatenated to perform low-light image enhancement guided by arbitrary references.

\noindent 2)
Our network achieves multi-level enhancement mapping trained with paired images. In the training datasets, each low-light image only has one corresponding normal-light image. By comparison, existing methods trained with such datasets simply produce a one-to-one result.

\noindent 3)
Extensive experiments demonstrate strong capacity on various datasets. Furthermore, the network offers diverse outputs according to different brightness references.

\vspace{-5pt}
\section{Methodology}
\label{sec:methodology}

The goal of low-image enhancement is to learn a mapping from an image to a normal-light version. However, the NORMAL light level is within a range rather than a discrete value from both qualitative and quantitative point of view. Thus, it is suggested the enhancement is a one-to-many mapping given application scenarios or users' aesthetic. To achieve multi-level low-light image enhancement, we make basic assumptions in Sec.\ref{sec:assumptions}. Then the network structure and loss function used to optimize the network are described in detail in Sec.\ref{sec:architecture} and Sec.\ref{sec:loss function} respectively.

\vspace{-5pt}
\subsection{Assumptions}
\label{sec:assumptions}

\noindent\textbf{Assumption 1} An image can be decomposed into two feature components in the latent space, namely the content component and the luminance component.

Let $\vec{x}=\{x_{1},x_{2},\ldots,x_{n}\}$ be a set of images with different light levels in the same scene. For each image $x_i$, $f_i$ is its feature vector in the latent space, which consists of a content component $c$ and a luminance component $l_i$. In our assumptions, $c$ is invariant for light levels $i$, while $l_i$ is specific for $i$. In other words, a pair of corresponding images $(x_i,x_j)$, where $i\neq j$, are encoded by an encoder $E$ to generate feature vectors $f_i=E(x_i)$ and $f_j=E(x_j)$. In the latent space, $f_i$ and $f_j$ are decomposed into $(c_i,l_i)$ and $(c_j,l_j)$. Next, $c_i$ and $l_j$ are concatenated to form a new feature vector $f_{i}^{'}$, then $f_{i}^{'}=f_j$. The reconstructed image of $f_{i}^{'}$ by a decoder $G$ is the same as $x_j$. In this way, multi-level mapping is performed by extracting luminance components from images with diverse light levels.

\vspace{3pt}
\noindent\textbf{Assumption 2} Two feature components with fixed dimensions are low-coupling.

\begin{figure}[t]
	\centering
	\includegraphics[width=\linewidth]{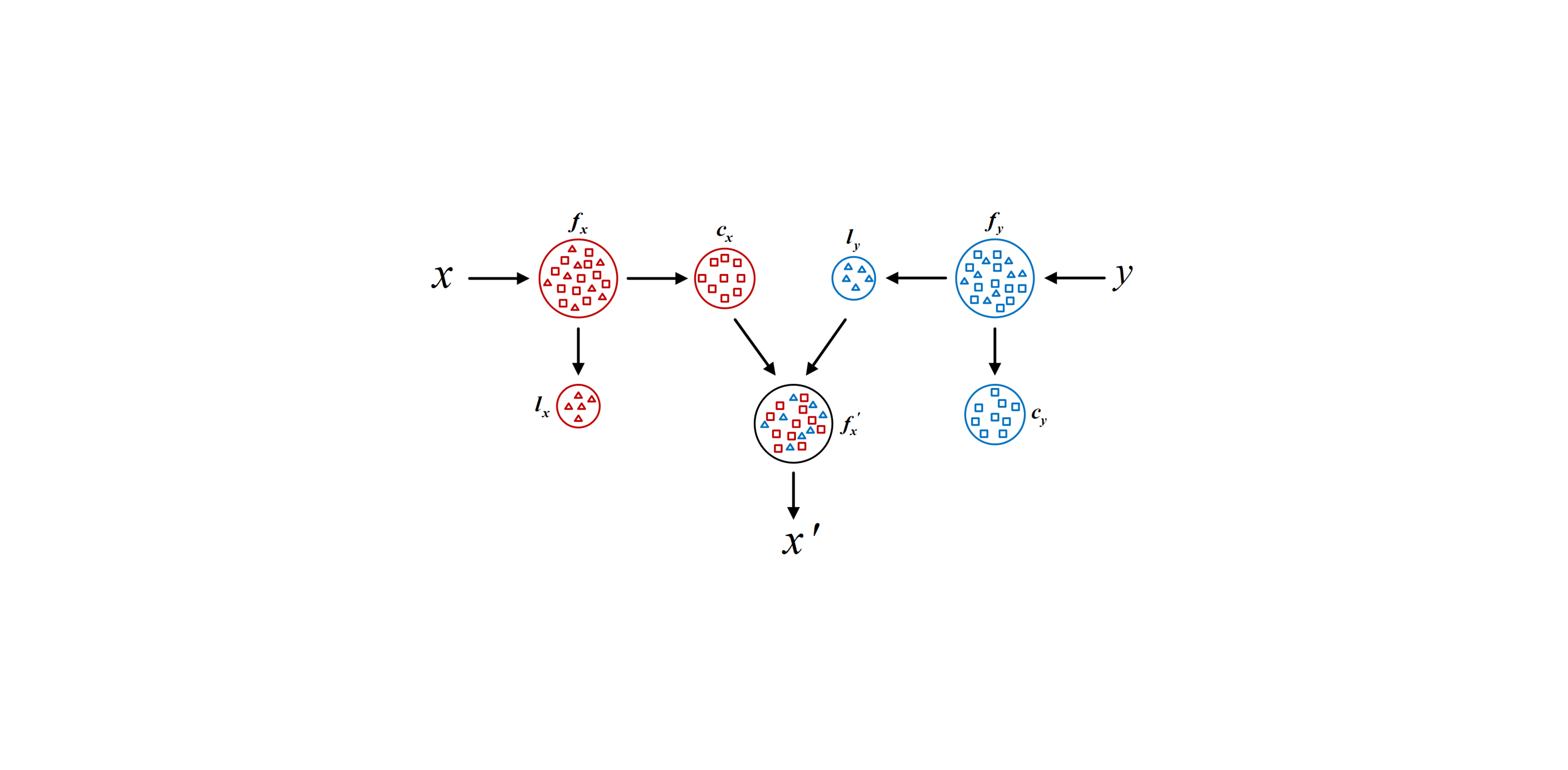}
	\vspace{-0.3cm}
	\caption[]{A schematic disgram of assumption 2. The image pair $(x,y)$ with irrelevant content is decomposed and concatenated in the latent space to generate an enhancement result of $x$.}
	\label{fig:assumptions}
	\vspace{-0.4cm}
\end{figure}

The above $\vec{x}$ is challenging to acquire in practice, so it is considered to use images that are content-irrelevant with the low-light images as a guideline. \textbf{\emph{This paper executes the multi-level low-light image enhancement task guided by arbitrary images as brightness references regardless of scenes.}} Thus, the components are expected to be low-coupling, so as to concatenate two images without involving information independent of brightness in the reference image. As shown in Fig.\ref{fig:assumptions}, let $(x,y)$ be an image pair with different scenes, where $y$ is an image as brightness reference. The goal of the task is learning a mapping from x to a corresponding version $x^{'}$ which is as bright as $y$. Specifically, the feature vectors of $x$ and $y$ are decomposed into $(c_x,l_x)$ and $(c_y,l_y)$ respectively, and then $c_x$ and $l_y$ are concatenated to reconstruct an enhancement result of $x$. The result preserves original scene-invariant information of $x$ and introduces target brightness from $y$. By taking different reference images as guidance, multi-level low-light image enhancement is achieved. The key to testify the assumptions is learning an encoder $E$ and a decoder $G$ using convolutional neural networks.

\vspace{-5pt}
\subsection{Architecture}
\label{sec:architecture}

\begin{figure*}[t]
	\vspace{0.1cm}
	\centering
	\includegraphics[width=\linewidth]{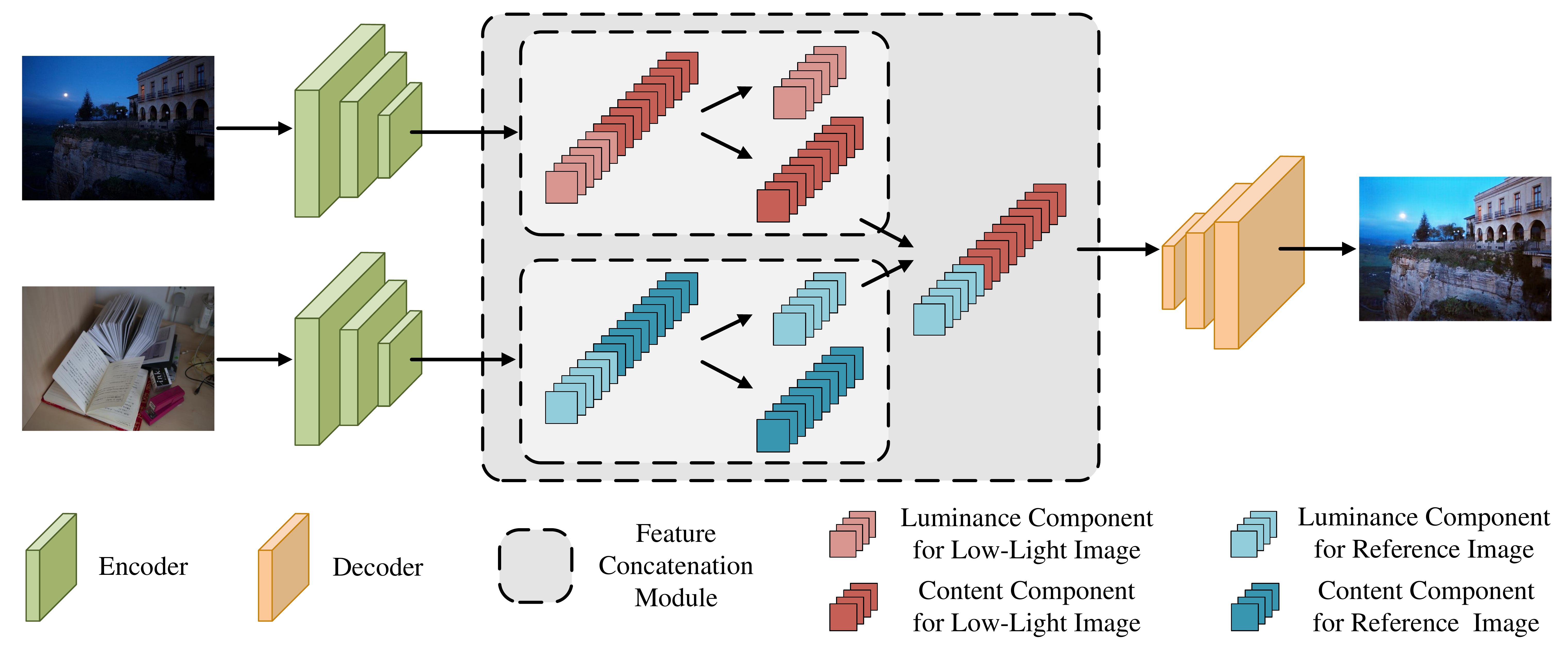}
	\vspace{-0.5cm}
	\caption[]{The architecture of the proposed method. It consists of an encoder $E$, a feature concatenation module and a decoder $G$. The same $E$ is used for two inputs.}
	\label{fig:architecture}
	\vspace{-0.4cm}
\end{figure*}

Our model is designed to enhance a low-light image to corresponding normal-light versions. We present the network structure in Fig.\ref{fig:architecture}. It consists of an encoder $E$, a feature concatenation module and a decoder $G$, which form a U shape. The network takes two images as input, including a low-light image $I_i$ and a reference image $I_r$. During training, $I_r$ and $I_i$ are identical in content, while in testing, $I_r$ is an arbitrary image. The same $E$ is used for both inputs. 

Our network employs down-sampling part of U-Net\cite{Unet} as the encoder $E$, followed by a global average pooling, which respectively encodes $I_i$ and $I_r$ as feature vectors $f_i$ and $f_r$. Correspondingly, the decoder $G$ is up-sampling part of U-Net to reconstruct the feature vector. Details about the feature concatenation module are then provided, which is a crucial part of our network.

\vspace{3pt}
\noindent\textbf{The Feature Concatenation Module}

Its function is to regroup components from two input feature vectors, so that the output vector contains all desired information. Specifically, $f_i$ and $f_r$ are fed into the feature concatenation module, and their components are concatenated to obtain a new feature, which consists of $c_i$ and $l_r$. Finally, the model produces the concatenation feature map through a fully connection layer and dimension expansion operation, which has the same resolution and channels as corresponding feature map in the encoding stage.

The low-light image is enhanced by introducing $l_r$ while retaining $c_i$. This way alleviates the problem of color distortion and accords with essence of the task, that is, only light level changes.

As stated in the assumptions, input feature vectors are decomposable, and decomposed components are low-coupling. Therefore, the proposed method uses loss functions described in Sec.\ref{sec:loss function} to limit fixed dimensions of the vectors to include brightness information alone, and remaining dimensions include other information such as color, structure and details. These two kinds of information are non-overlapping.

\vspace{-5pt}
\subsection{Loss Function}
\label{sec:loss function}

To perform the task, we propose several differentiable losses to restrict image-to-image and feature-to-feature processes. The following three components of losses are minimized to train our network. 

\vspace{3pt}
\noindent\textbf{Reconstruction loss } 
In the image-to-image process, we compute the reconstruction loss. The $L_1$ error is used to measure distance between the prediction and the ground truth. The reconstruction loss can be expressed as:
\begin{equation}
L_r={\parallel G(c_i,l_r)-I_r\parallel}_1
\end{equation}

where $I_i$ and $I_r$ are respectively low-light and reference normal-light images, $c_i$ is the content component decomposed by $I_i$, and $l_r$ is the luminance component decomposed by $I_r$. Pixels of all channels in the inputs of the network are normalized to $[0,1]$.

The loss ensures that the network decomposes image pairs $(I_i,I_r)$ with the same content into identical content components and different luminance components, which is achieved by reconstructing the feature vector composed of $c_i$ and $l_r$ into an image consistent with $I_r$.

\vspace{3pt}
\noindent\textbf{Feature loss}
The feature loss is designed for feature-to-feature mapping. It is expected that feature components can be reconstructed after passing through the decoder and encoder. To this end, we use content feature loss and luminance feature loss to constrain and learn reconstruction and extraction processes of feature components. The feature loss is expressed as:
\begin{equation}
L_f=L_{f\_c}+L_{f\_l}
\end{equation}

Here, $L_{f\_c}$ and $L_{f\_l}$ are respectively the content feature loss and the luminance feature loss. Specifically, the content feature loss is defined as:
\begin{equation}
L_{f\_c}={\parallel c_p-c_i\parallel}_2
\end{equation}

where $c_i$ and $c_p$ represent the content components of the low-light image and the prediction. ${\parallel\cdot\parallel}_2$ is the $L_2$ error. The content feature loss, on the one hand, ensures that the content component is unchanged after enhancement, and on the other hand encourages the feature to be consistent with the original after decoding and encoding. Next, we refer to the definition of the triplet loss to define the luminance feature loss as:
\begin{equation}
L_{f\_l}={[\mathcal{D}(l_p,l_r)-\mathcal{D}(l_p,l_i)+\alpha]}_+
\end{equation}

where $l_i$, $l_r$, and $l_p$ respectively represent the luminance components of the low-light image, reference image, and the prediction. ${[\cdot]}_+$ is a rectifier. The loss is the value in the rectifier when it is greater than 0; otherwise, the loss is 0. $\mathcal{D}(\cdot)$ is the squared Euclidean distance between feature vectors. $\alpha$ is a margin and is set to 0.08 by taking average distances between the luminance components of 20 image pairs, which are randomly selected from the dataset.

\begin{table*}[ht]
	\begin{center}
		\vspace{-0.4cm}
		\caption{Low-light image enhancement evaluation on the LoL dataset. The best result is bolded for PSNR and SSIM.}
		\vspace{0.1cm}
		\begin{tabular}{c|cccccccccccccc}
			\toprule
			& & & & & & & & & & &\\[-8pt]
			Method         & CRM   & Dong  & LIME  & MF    & Retinex-Net & MSR   & NPE   & GLAD  & KinD  & MIRNet & Ours  \\
			& \cite{2017A}   & \cite{2011Fast}  & \cite{Guo2017}  & \cite{2016A}    & \cite{Chen2018} & \cite{2017MSR}   & \cite{wang2013naturalness}   & \cite{wang2018gladnet}  & \cite{zhang2019kindling}  & \cite{Zamir2020MIRNet} &\\\midrule
			& & & & & & & & & & &\\[-8pt]
			PSNR$\uparrow$ & 17.20 & 16.72 & 16.76 & 18.79 & 16.77       & 13.17 & 16.97 & 19.72 & 20.87 & 24.14  & \textbf{27.90} \\
			& & & & & & & & & & &\\[-8pt]
			SSIM$\uparrow$ & 0.64  & 0.58  & 0.56  & 0.64  & 0.56        & 0.48  & 0.59  & 0.70  & 0.80  & 0.83   & \textbf{0.86}  \\ \bottomrule
		\end{tabular}
		\label{tab:compare_lol}
	\end{center}
	\vspace{-0.5cm}
\end{table*}

We choose triplet form rather than the $L_2$ metric used in the content feature loss. The reason is that $l_p$ is expected to be similar to $l_r$ and different from $l_i$ on account of specificity of the luminance component.

\vspace{3pt}
\noindent\textbf{Content consistency loss}
Next, the content consistency loss is employed to restrict the enhanced image to be the same as the original low-light image except for the light level. Images are first mapped to the HSV color space. Optimization process penalizes the cosine distance of H and S channels between the prediction and the low-light image. The content consistency loss is expressed as:
\begin{equation}
L_c=L_{c\_H}+L_{c\_S}
\end{equation}

Here, $L_{c\_H}$ and $L_{c\_S}$ respectively represent the cosine loss of H and S channel expressed as:
\begin{equation}
L_{c\_H}=1-\angle(H_p,H_i)
\end{equation}

\begin{equation}
L_{c\_S}=1-\angle(S_p,S_i)
\end{equation}

where $\angle(,)$ is an operation to calculate cosine similarity. $H_i$ and $H_p$ are the H channel of the low-light image and prediction, respectively. Similarly, $S_i$ and $S_p$ are the S channel.

We support such color space mapping based on the following experiments. If the H and S channels of the low-light image are combined with the V channel of the content matching normal-light image, and then mapped back to the RGB space, the result is nearly the same as the normal-light image. It proves that the similarity of the H and S channels between the prediction and the low-light image is able to measure whether scene-invariant information changes after enhancement.

Cosine loss is adopted instead of the $L_1$ loss for the following reasons. First, the $L_1$ metric has been calculated in the RGB color space, which fails to figures whether the directions of pixel values are the same. In addition, it is experimentally observed that the enhanced image color is closer to the ground truth when using the cosine loss compared to the $L_1$ loss.

\vspace{3pt}
\noindent\textbf{Total loss}
The proposed network is optimized using the total loss:
\begin{equation}
L_total=L_r+\lambda L_f+L_c
\end{equation}

where $\lambda$ is a weight of corresponding loss term.

\section{Experiments}

In this section, we begin with dataset and implementation details for training. The proposed method is compared with state-of-the-art methods according to extensive qualitative and quantitative experiments. Moreover, the ability to generate multi-level enhancement results is demonstrated with arbitrary brightness references.

\vspace{3pt}
\noindent\textbf{Dataset} 

LoL dataset\cite{Chen2018} is involved in training. It consists of 500 image pairs, where each pair contains a low-light image and its corresponding normal-light image. The first 485 image pairs are for training and the remaining are for testing.

\vspace{3pt}
\noindent\textbf{Implementation Details} 

Our network is implemented with Tensorflow on an NVIDIA 2080Ti GPU. The model is optimized using Adam optimizer with a fixed learning rate of 1e-4. Batch size is set to 8. We train the model for 1000 epochs with the whole image as input. For data augmentation, a horizontal or vertical flip is randomly performed. Besides, a 100$\times$100 image patch is stochastically located from each low-light image and is replaced with an image patch at the same position in the ground truth. The weight $\lambda$ is set to 2.

The network is trained in an end-to-end manner. During the training, a low-light and a reference image are taken as input. After passing through our model, an enhanced image is generated, which is also fed into the encoder. The feature concatenation module produces feature components of three images to calculate the feature loss. 

\vspace{-5pt}
\subsection{Performance Evaluation}

Our method is evaluated on widely used datasets, including the LoL, LIME\cite{Guo2017}, DICM\cite{lee2012contrast}, NPE\cite{wang2013naturalness} and MEF\cite{ma2015perceptual} datasets. The effectiveness of the proposed algorithm is demonstrated by qualitative and quantitative comparison with several state-of-the-art methods, such as KinD\cite{zhang2019kindling}, MIRNet\cite{Zamir2020MIRNet} and PieNet\cite{kim2020pienet}. For PieNet, only numerical result published by the author is used for quantitative comparisons since source code is non-available. For the LIME, NPE and MEF datasets, we merely conduct qualitative experiments due to the lack of the ground truth.

\begin{figure*}
	\begin{center}
		\begin{tabular}{c@{ }c@{ }c@{ }c@{ }c@{ }}
			\includegraphics[width=0.24\linewidth]{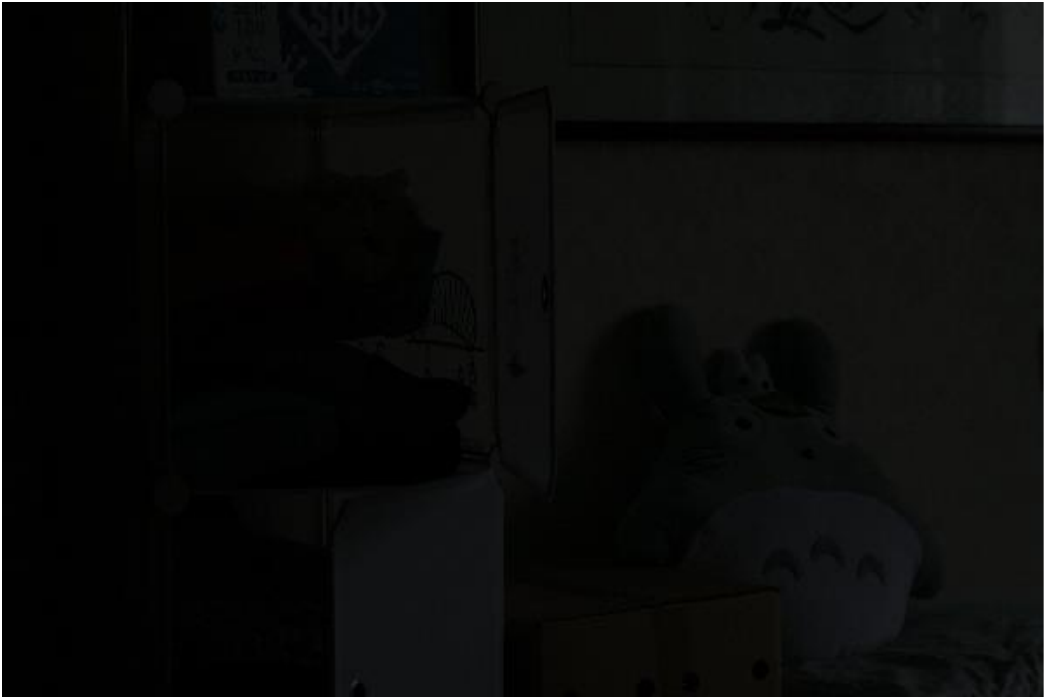}&
			\includegraphics[width=0.24\linewidth]{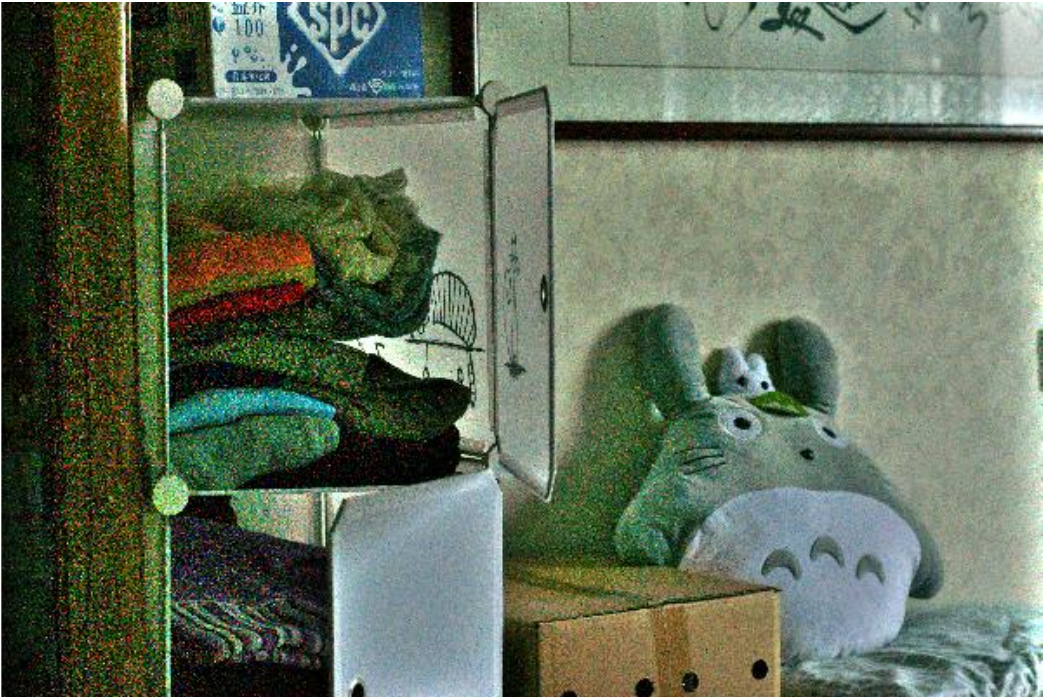}&
			\includegraphics[width=0.24\linewidth]{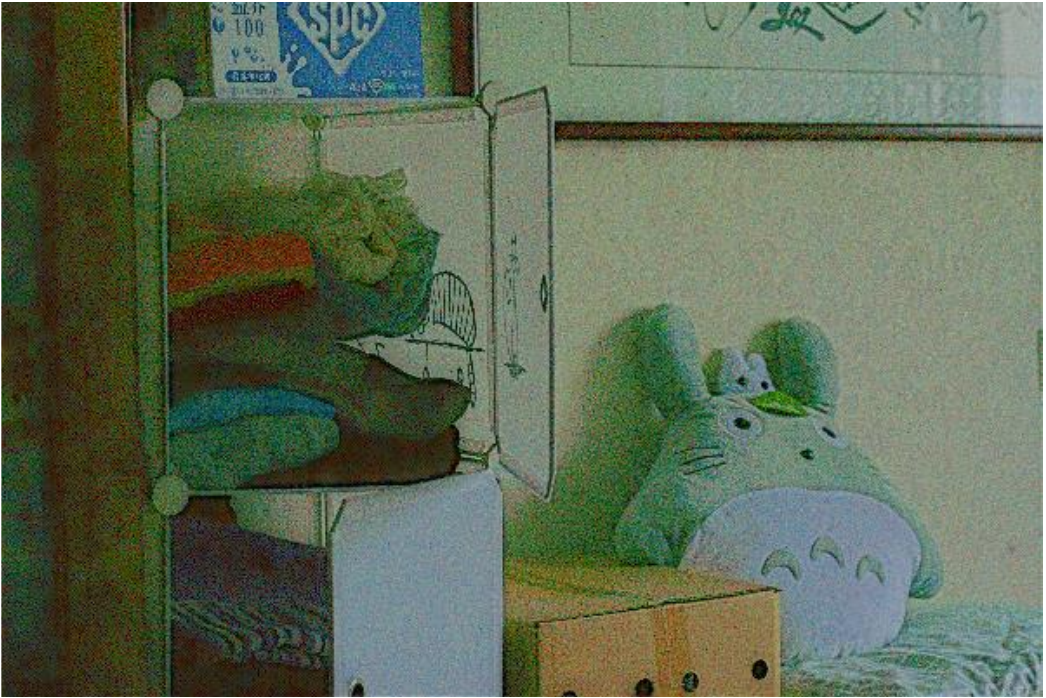}&
			\includegraphics[width=0.24\linewidth]{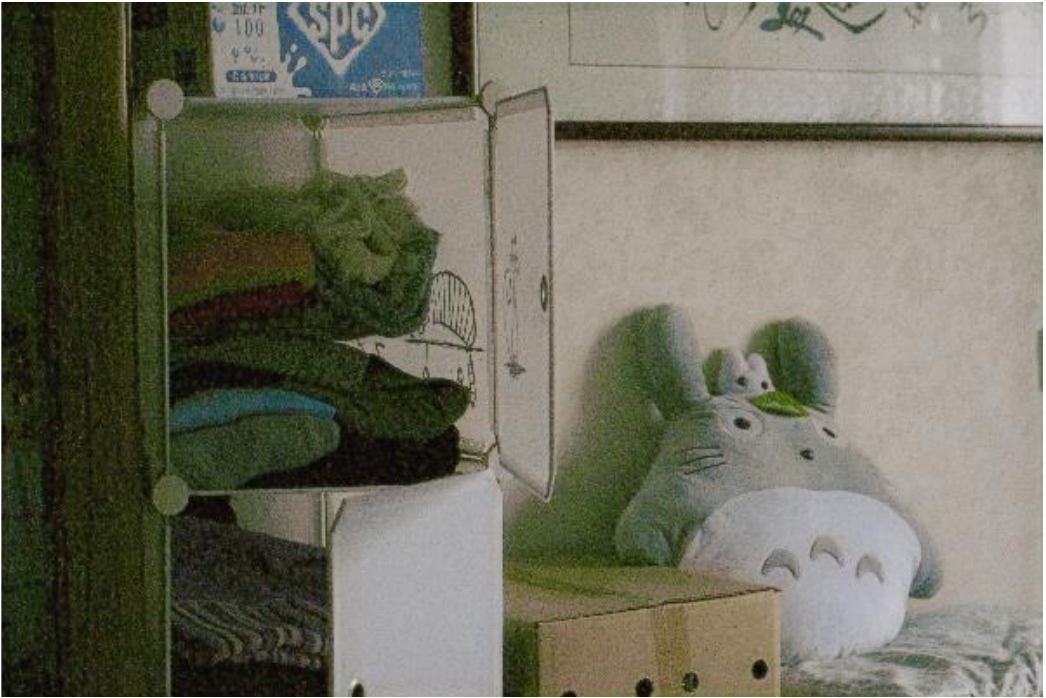}\\
			(a) Input & (b) LIME\cite{Guo2017} & (c) Retinex-Net\cite{Chen2018} & (d) GLAD\cite{wang2018gladnet}\\
			\includegraphics[width=0.24\linewidth]{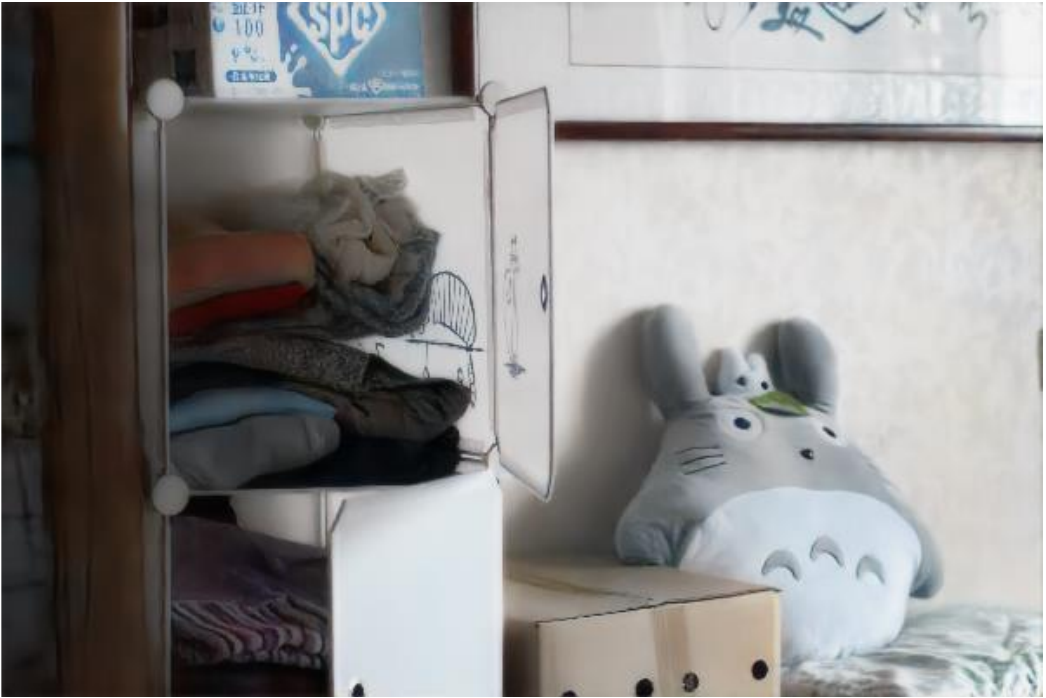}&
			\includegraphics[width=0.24\linewidth]{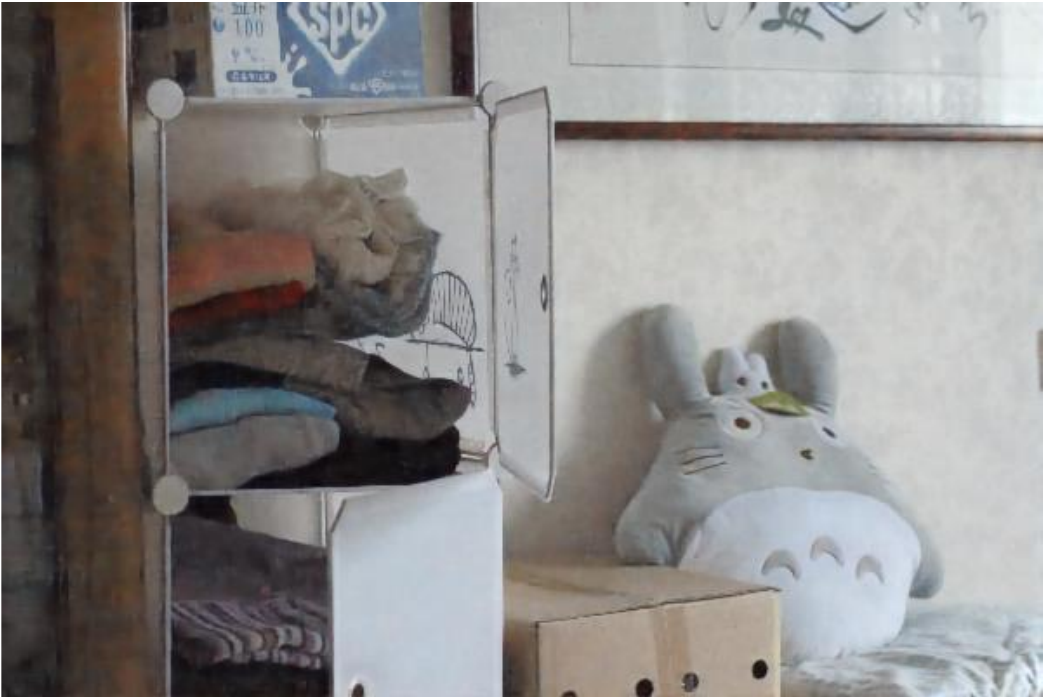}&
			\includegraphics[width=0.24\linewidth]{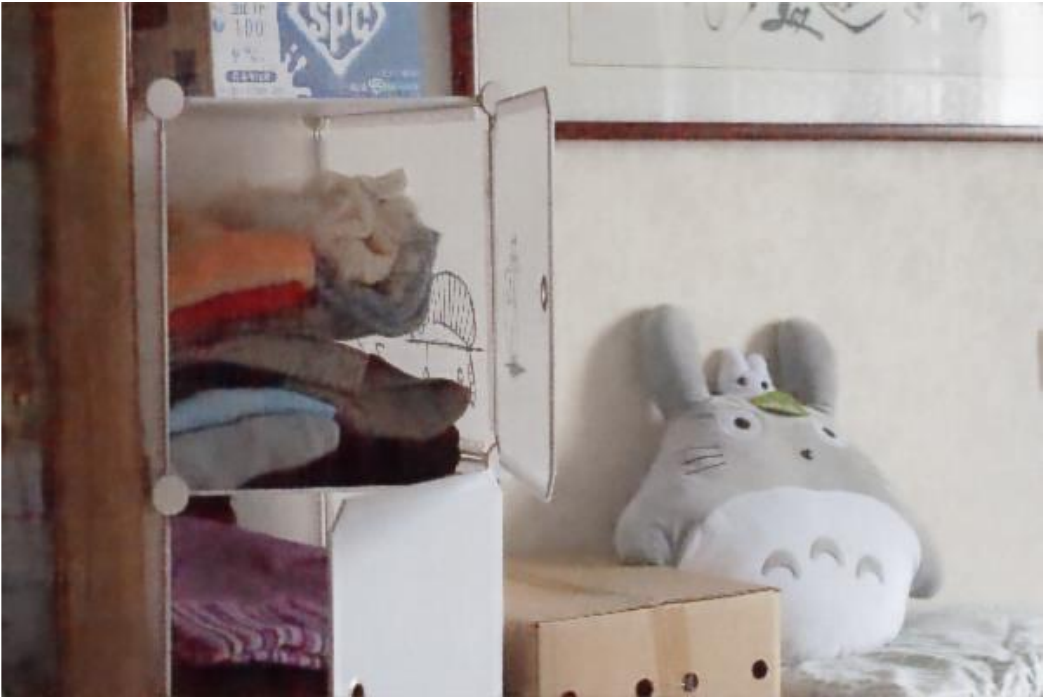}&
			\includegraphics[width=0.24\linewidth]{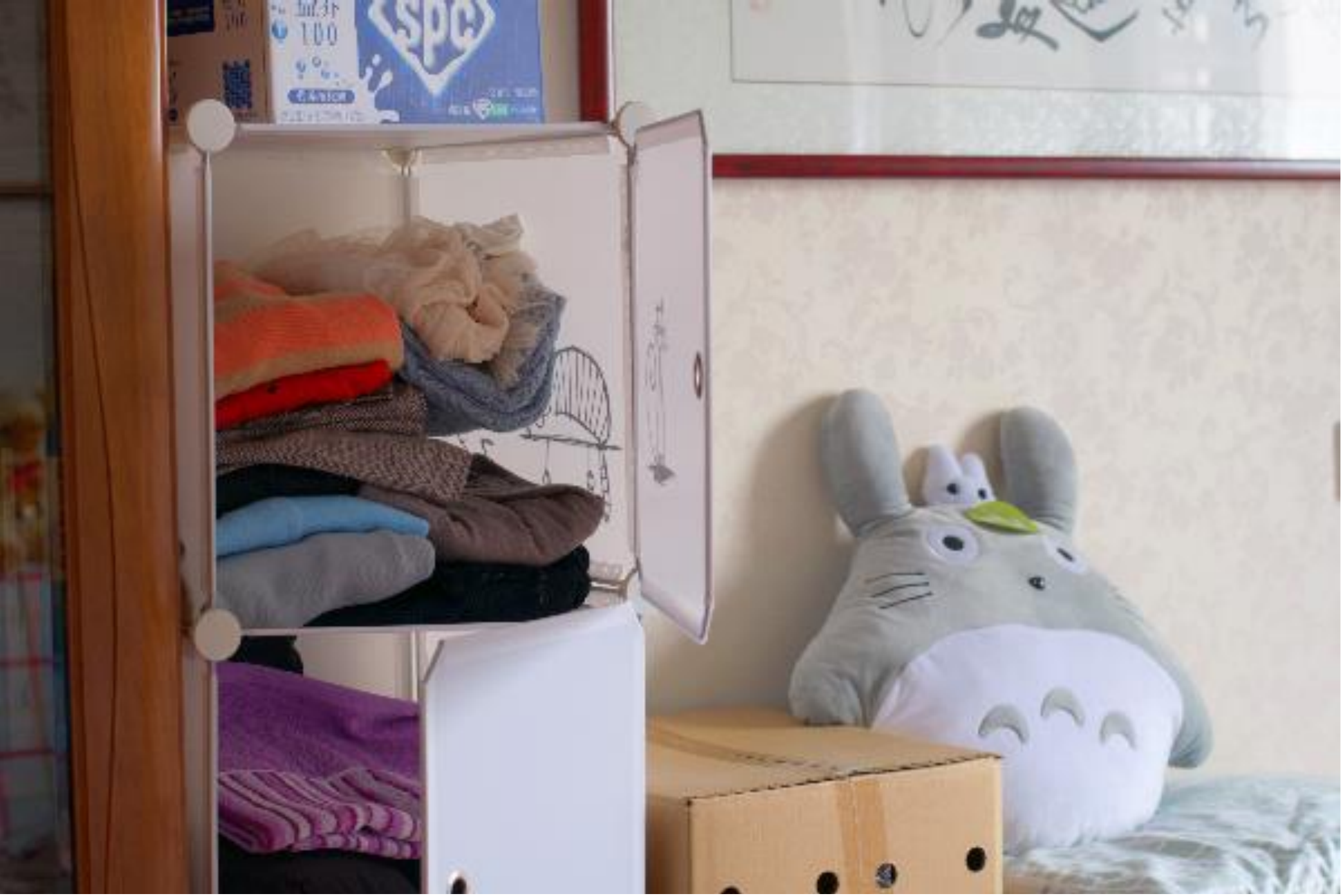}\\
			(e) KinD\cite{zhang2019kindling} & (f) MIRNet\cite{Zamir2020MIRNet} & (g) Ours & (h) Ground Truth\\
		\end{tabular}
	\end{center}
	\vspace{-0.5cm}
	\caption{Visual comparison with ground truth. Our result is more satisfactory than others in terms of brightness and colors.}
	\label{fig:with_gt}
	\vspace{0.2cm}
\end{figure*}

\begin{figure*}[t]
	\begin{center}
		\begin{tabular}{c@{ }c@{ }c@{ }c@{ }c@{ }c}
			\includegraphics[width=.19\textwidth]{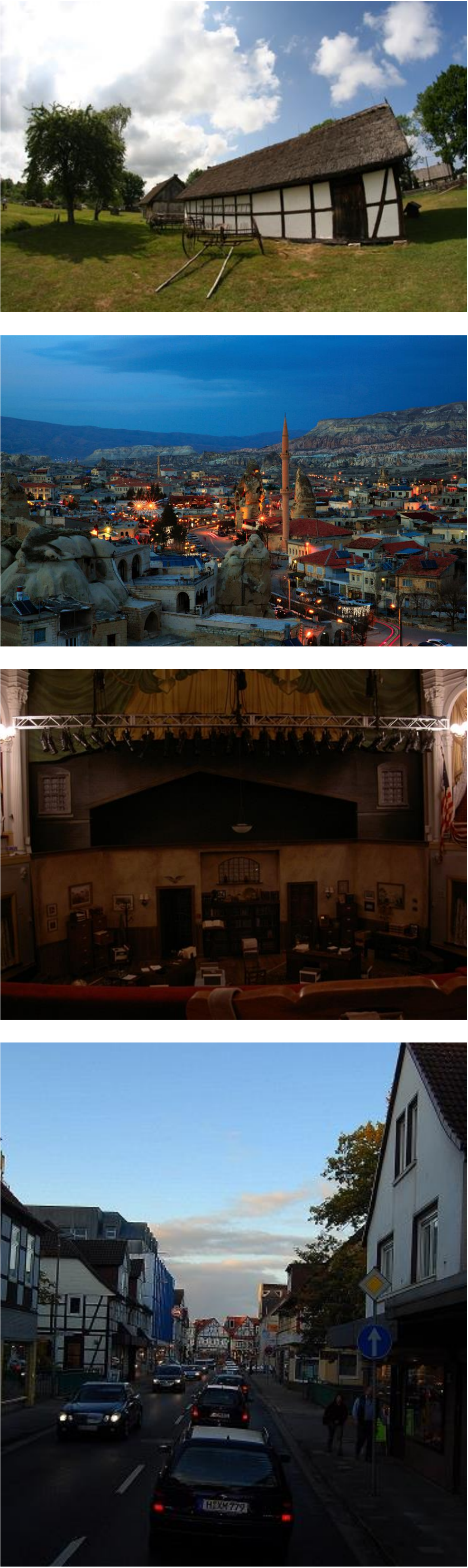}~&
			\includegraphics[width=.19\textwidth]{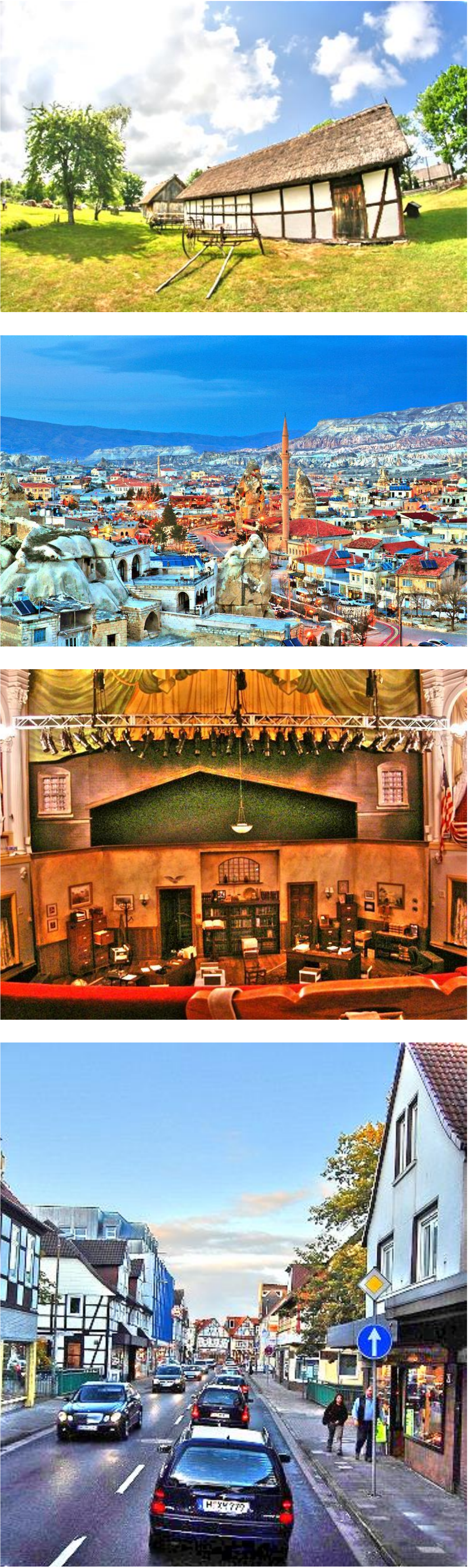}~&
			\includegraphics[width=.19\textwidth]{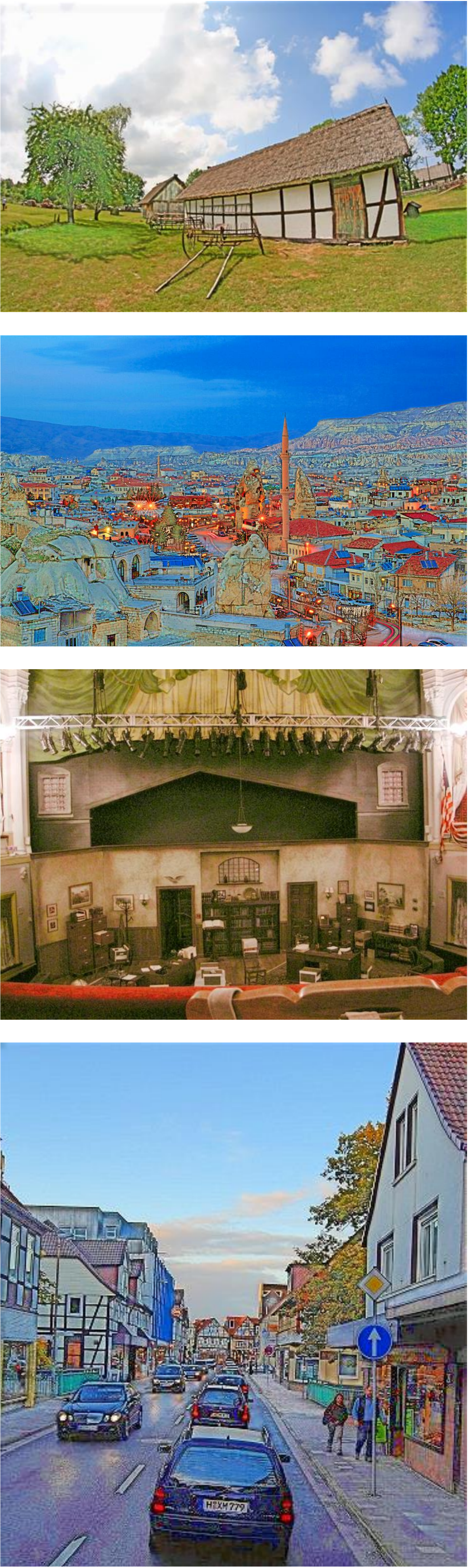}~&
			\includegraphics[width=.19\textwidth]{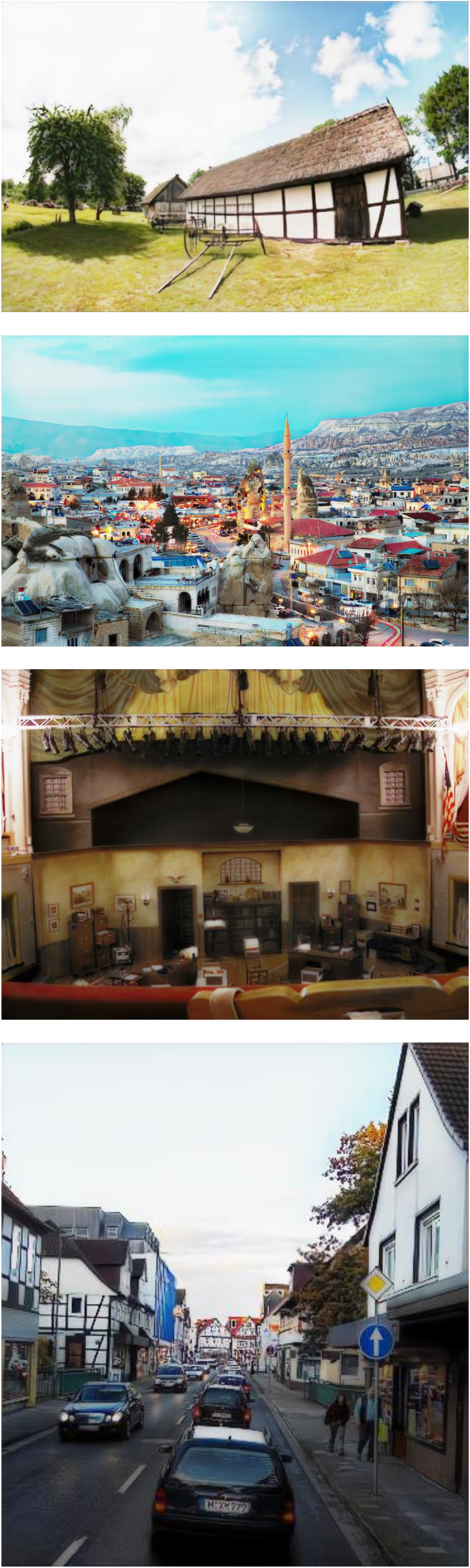}~&
			\includegraphics[width=.19\textwidth]{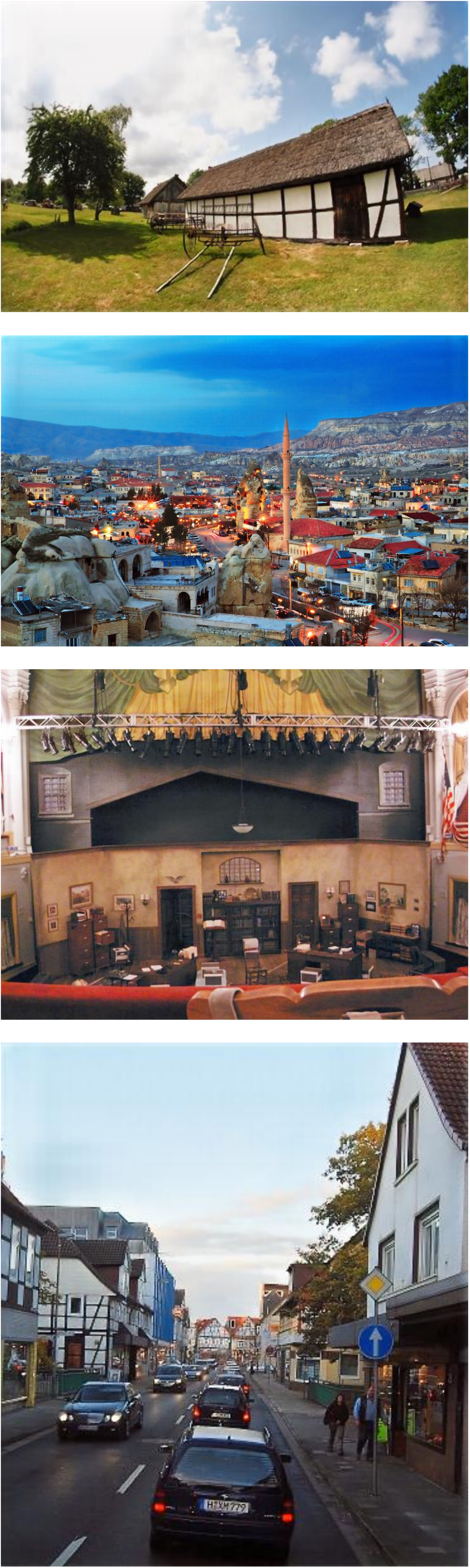}\\
			(a) Input~& (b) LIME~\cite{Guo2017}& (c) Retinex-Net~\cite{Chen2018}& (d) KinD~\cite{zhang2019kindling}& (e) Ours\\
		\end{tabular}
	\end{center}
	\vspace{-0.5cm}
	\caption{Visual comparison without ground truth. The first column images are respectively from the MEF, NPE, DICM and LIME datasets and under different lighting conditions.}
	\vspace{-0.2cm}
	\label{fig:without_gt}
\end{figure*}

\begin{figure*}[t]
	\centering
	\includegraphics[width=0.9\linewidth]{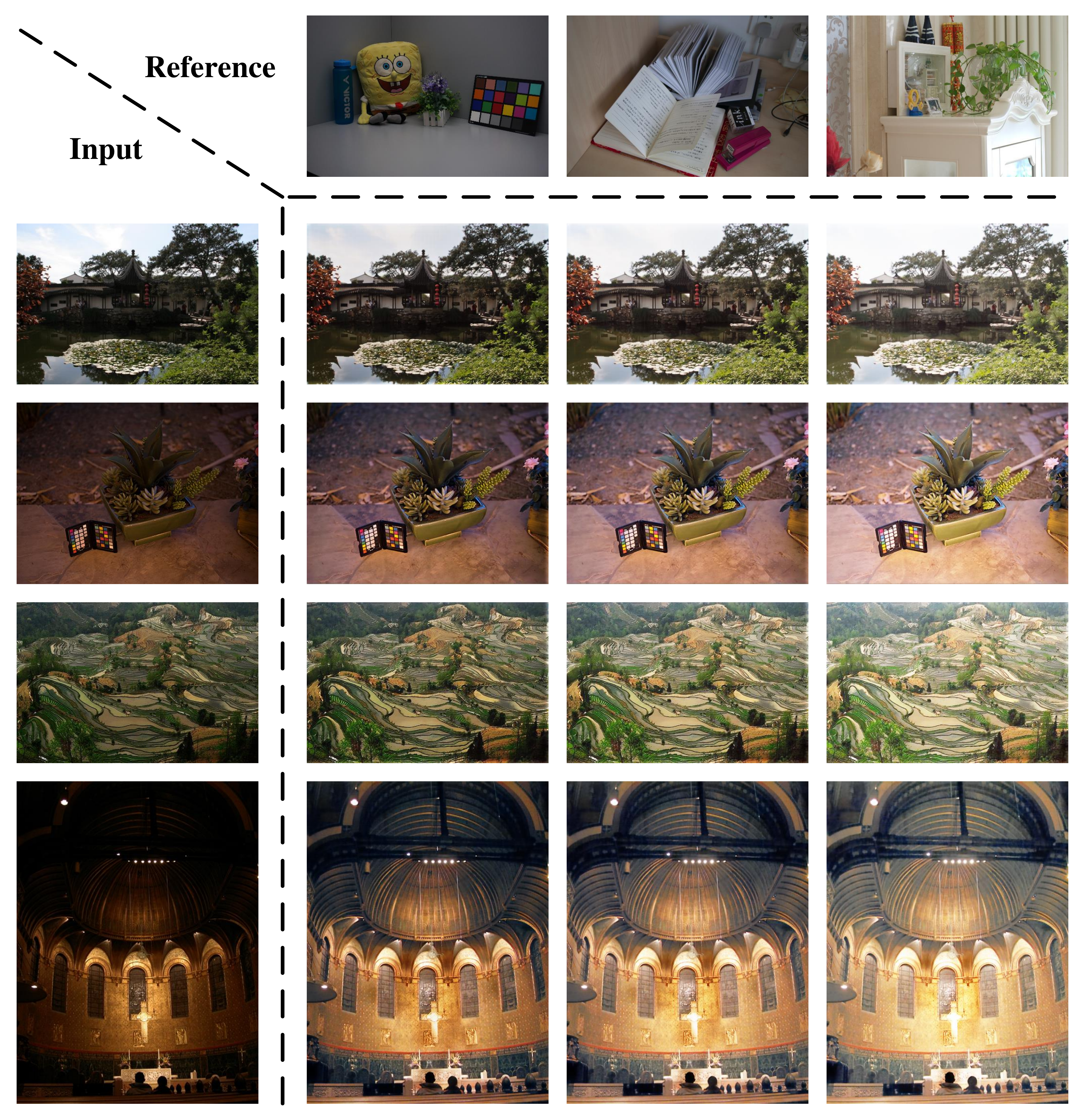}
	\caption{Muti-level enhancement results. The ability of our network to generate multiple enhancement versions for the same low-light image is demonstrated.}
	\label{fig:multi-level}
	\vspace{-0.5cm}
\end{figure*}

\vspace{-5pt}
\subsubsection{Quantitative Comparison}

In quantitative comparison, PSNR and SSIM\cite{wang2004image} are calculated as evaluation metrics. Generally, high value indicates better results. For a fair comparison, we compare the proposed model with methods trained on the same data. Furthermore, methods involved in the comparison all employ the default training set and test set. Table\ref{tab:compare_lol} reports PSNR/SSIM results of our method and several others on the LoL dataset. The best result is bolded for each metric. As we can see from the table that our network significantly outperforms all the other methods. Notably, the proposed model achieves 3.76dB better than MIRNet on the LoL dataset, which is currently optimal. There are two main reasons. First, the way of feature concatenation retains the scene-invariant information of the low-light image to the greatest extent, which alleviates color distortion. Second, well designed loss functions improve the performance of our network.

\vspace{-5pt}
\subsubsection{Visual and Perceptual Comparisons}

Figures \ref{fig:with_gt} and \ref{fig:without_gt} give visual comparisons on low-light images from five datasets, which are under different lighting conditions. As shown in Fig.\ref{fig:with_gt}, by comparing with the ground truth, our method not only enhances dark regions but also makes colors of the enhanced image closer to the ground truth. In the absence of ground truth, as can be seen from results of different methods shown in Fig.\ref{fig:without_gt}, our method is more natural in appearance, making images look more realistic. In contrast, other methods either fail to enhance images or suffer from more or less degradations, such as noise and color distortion. In a word, the proposed method achieves better contrast, more vivid colors and sharper details, which are more satisfying.

\subsection{Different level of Enhancement}

We show results of multi-level mapping in Fig.\ref{fig:multi-level}. Our network is able to generate multiple enhancement versions for the same low-light image guided by various reference images. More importantly, versions enhanced by an image are basically the same in details, structures and colors except for light levels. In addition, when different low-light images are matched with the same reference image, results have approximate brightness.

Most existing methods trained on paired datasets simply generate one fixed brightness result for a low-light image, which is a one-to-one mapping and means the lack of diversity. In contrast, our method achieves multi-level enhancement utilizing such datasets.

\vspace{-8pt}
\section{Conclusion}
\vspace{-6pt}
In this paper, we focus on the subjectivity of the enhancement and introduce brightness reference to produce results that conform to this property. We propose a deep network for multi-level low-light image enhancement guided by a reference image. In the network, an image is decomposed into two low-coupling feature components in the latent space, and then the content and luminance component of two images are concatenated to generate a new image. Multiple normal-light versions of one low-light image are obtained by selecting different reference images as guidelines. Extensive experiments demonstrate the superiority of our method compared with existing state-of-the-art methods.

{\small
\bibliographystyle{ieee_fullname}
\bibliography{egbib}
}

\end{document}